\documentclass[conference]{IEEEtran}
\ifCLASSINFOpdf
\else
\fi
\usepackage{comment}
\usepackage{color}
\usepackage{tabularx}
\usepackage{graphicx}
\usepackage{pdfpages}

\usepackage{float}
\usepackage[numbers]{natbib}

\begin{document}
%
\title{Feature-Level Fusion of Super-App and Telecommunication Alternative Data Sources for Credit Card Fraud Detection}

\makeatletter
\newcommand{\linebreakand}{%
  \end{@IEEEauthorhalign}
  \hfill\mbox{}\par
  \mbox{}\hfill\begin{@IEEEauthorhalign}
}
\makeatother

\author{\IEEEauthorblockN{Jaime D. Acevedo-Viloria}
\IEEEauthorblockA{Rappi Colombia\\
jaime.acevedo@rappi.com}
\and
\IEEEauthorblockN{Sebastián Soriano Pérez}
\IEEEauthorblockA{Rappi México\\
sebastian.soriano@rappi.com}
\and
\IEEEauthorblockN{Jesus Solano}
\IEEEauthorblockA{Rappi Colombia\\
jesus.solano@rappi.com}
\linebreakand
\IEEEauthorblockN{David Zarruk-Valencia}
\IEEEauthorblockA{Rappi Colombia\\
davidzarruk@rappi.com}
\and
\IEEEauthorblockN{Fernando G. Paulin}
\IEEEauthorblockA{Rappi México\\
fernando.gonzalez@rappi.com}
\and
\IEEEauthorblockN{Alejandro Correa-Bahnsen}
\IEEEauthorblockA{Rappi Colombia\\
alejandro.correa@rappi.com}}


%


\maketitle

\begin{abstract}
Identity theft is a major problem for credit lenders when there’s not enough data to corroborate a customer’s identity. Among super-apps—large digital platforms that encompass many different services—this problem is even more relevant; losing a client in one branch can often mean losing them in other services. In this paper, we review the effectiveness of a feature-level fusion of super-app customer information, mobile phone line data, and traditional credit risk variables for the early detection of identity theft credit card fraud. Through the proposed framework, we achieved better performance when using a model whose input is a fusion of alternative data and traditional credit bureau data, achieving a ROC AUC score of 0.81. We evaluate our approach over approximately 90,000 users from a credit lender's digital platform database. The evaluation was performed using not only traditional ML metrics but the financial costs as well.
\end{abstract}


%
\IEEEpeerreviewmaketitle

\section{Introduction}
Fraud detection is of utmost importance for credit lending companies. Fraud events not only generate financial losses to the lender, that according to the 2020 ACFE report \cite{2020acfe} amounted up to $5\%$ of revenues each year, but they also generate losses in terms of a negative impact to the credibility and reputation of the company. According to One World Identity \cite{2020owi}, digital fraud rates have also accelerated in 2020 with the increase in digital transactions due to the pandemic and, in spite of this, regulations are still slow when trying to catch up to the fraudulent actors' tactics.

Current commonly utilized fraud detection models involve either machine learning models or hard rule-based systems that mainly use available credit-bureau data or the users' credit history as inputs for the detection of fraudulent behaviour. Models that use traditional data can be less effective at detecting fraud since the fraudsters' strategies constantly evolve as they learn the patterns in the data that are associated to fraudulent behaviour by the ML model or the rules-based system. 

On the other hand, super-apps are digital platforms that encompass many different services that could profile a user. Some services offered by super-apps include but are not restricted to financial services, food delivery services, marketplaces, travel shopping, and lodging services. Super-apps use the collected interactions to enhance the performance of models in several fields~\cite{liu2017, wang2019, roa2021super, acevedoviloria2021}. Such models have gained a lot of traction in the market with companies such as WeChat and AliPay that have shown promising results in profiling financial behavior patterns without traditional financial data~\cite{liu2017}. Furthermore, telecommunications data includes features such as the type of phone line the user has or the monthly average amount spent on the phone line, among others. This data is not generated within the super-app, therefore we study whether it contributes to the fraud detection models performance or not. This not only allows us to add more information to our prediction systems, but it also helps leveling the playing field for previously non-banked or under-banked consumers \cite{2020owi}.



Consequently, it could be interesting to explore how much alternative data (such as consumption patterns, registered addresses, credit cards, devices, telecommunication signals, etc.) enhances a model's performance in the domain of credit card fraud detection. In this paper, we look at the effect of using super-app data as well as telecommunications data on credit fraud detection through feature-level fusion. Specifically, we look at the gains of creating features from additional sources of data, beyond the traditional credit bureau features. In this sense, our results apply to the set of users that are part of the super-app and also have a mobile phone line contract.

Given the inherent costs associated to fraud detection strategies faced by super-apps, as described above, we also measure the impact of our models in terms of the financial losses generated by the strategy, assigning specific costs to false positives and false negatives.

We propose a framework where we test a model using different sets of features and determine the advantages of each along with traditional credit bureau features. We compare each model in terms of how they improve the detection of identity theft at the origination of the credit line and what are the financial savings associated to them. The compared models' inputs are: super-app data only, phone-line data only, bureau data only, super-app $\&$ phone-line data, super-app $\&$ bureau data, and super-app $\&$ phone-line $\&$ bureau data all together; where we perform a feature-level fusion for each of the proposed combinations. Additionally, we complement these comparisons with the use of SHAP values \cite{lundberg2017} for the estimation of the features' importance in each of the models.

In summary, the main contribution of this paper are: 
\begin{itemize}
    \item A comprehensive evaluation of the impact of non-traditional data such super-app information and mobile phone contract data to enhance credit card fraud detection.
    \item A custom financial metric that includes a cost-sensitive approach to distinguish the cost of blocking a legitimate user and the cost of lending to a fraudster.
    \item A sensitivity analysis on the features that impact the fraud risk score the most using Shapley Additive explanation. 
\end{itemize}

The paper is organized as follows: Section~2 presents previous relevant work to this paper, Section~3 describes the proposed approach, Section~4 gives a thorough description of the used dataset along with the results and discussion, and finally, in Section~5 we present the conclusions of the paper.






\section{Background}

Fraud detection is a quite popular subject of research in the literature, with many different types of fraud being handled by different machine learning techniques \cite{awoyemi2017, varmedja2019, RYMANTUBB2018130}. Traditional machine learning models are commonly used in the industry, with models like logistic regression and decision trees taking a big part of the participation \cite{abdallah2016fraud}. 

Data mining techniques have also been found to be effective in the literature to combat Credit Card Fraud. For instance, Correa-Bahnsen et al.~\cite{bahnsen2016} proposed a new set of features based on analyzing the periodic behaviour of the time of a transaction, increasing their saving by 13\% in a fraud dataset provided by a card processing company. In the same vein, Veeramani et al.~\cite{veeramani2020} proposed an approach utilizing Isolation Forest and Local Outlier Factor for the detection of credit card payment fraud, achieving a F1-score of 0.28 when using the approach in a database of about 280,000 transactions.

Deep Learning techniques have also been leveraged for the detection of fraudulent credit card transactions. In Branco et al.~\cite{branco2020} the authors suggest interleaved sequence RNN's for the detection of fraudulent payments, in two different time-split databases they obtain improvements over a LightGBM baseline model with recall improvements of at least 3.2\%. Furthermore, these deep learning techniques have also been adapted for graphs for the improvement of fraud detection. Wang et al.~\cite{wang2019} proposed a Semi-supervised Graph Attentive Network for the detection of Financial Fraud, where they achieve a ROC Curve AUC of 0.807 that comfortably performed better than other compared models. More recently, Acevedo-Viloria et al.~\cite{acevedoviloria2021} proposed a framework of relational graph convolutional networks for the prevention of fraudulent behaviour, achieving an AUC of 0.80 on a time-split dataset containing around 50,000 users.

\subsection{Super-App Data}

As opposed to traditional lenders, super-apps have access to a wide range of services that provide additional information, both behavioural and transactional, from their users. This information allows them to create a better profile of their clients. With the evolving nature of fraudulent behaviour, where fraudsters may adapt to the current machine learning models leveraging traditional sources of data, the use of non-traditional information can provide an edge for the company in terms of the proper detection of fraud.

According to FICOblog \cite{2017fico}, alternative data for credit granting refers to any data that is not directly related to a consumer's credit behaviour. They also refer to traditional data as the credit bureau data of the lenders' own credit applications file. The authors also highlight how a combination of traditional and alternative data produces a more powerful model, reaching up to 15\% better performance when combining both data sources. As fraudsters adopt more complex tactics to adapt to the traditional data models, it is important to combine these features with alternative data to provide more robust fraud prevention solutions \cite{Hamilton2020}.

From these digital platforms we can better characterize the users in terms of their transactional behaviour, in services like food delivery, travel services, e-commerce purchases, transportation services, etc. Roa et al.~\cite{roa2021super} highlight how such variables give these platforms an advantage in terms of fraud prediction, using variables like: time of delivery consumption, amount spent in the digital platform, number of orders, tips as a percentage of total purchases, propensity to use discounts, among others; where they achieve an improvement of up to 2 percentage points of AUC. Additionally, the authors show how demographic data can be effectively leveraged from super-apps for fraud detection as well, adding information such as: the user's age, the locations from where they order, the device they use for the platform, the credit cards registered and used in the app, among others. Furthermore, Acevedo-Viloria et al.~\cite{acevedoviloria2021} show that not only are the variables taken from the a super-app environment useful for fraud detection, but also the interactions can be leveraged into graph models to further augment the performance of the classification task.

Another advantage of these types of data sources is that they don't require the user to have a credit history; therefore they lessen the barriers created for people that can't get into the traditional credit system. Super-apps allow us to leverage the many sources of information mentioned before to cover a higher spectrum of the population and properly assess them in terms of whether they may incur into origination fraud when asking for a credit line. This also has value in terms of opening up the market to people previously considered to be "credit invisible", which are common in developing markets \cite{2017fico}.


\begin{figure*}
        \centering
        \includegraphics[width=0.80\textwidth]{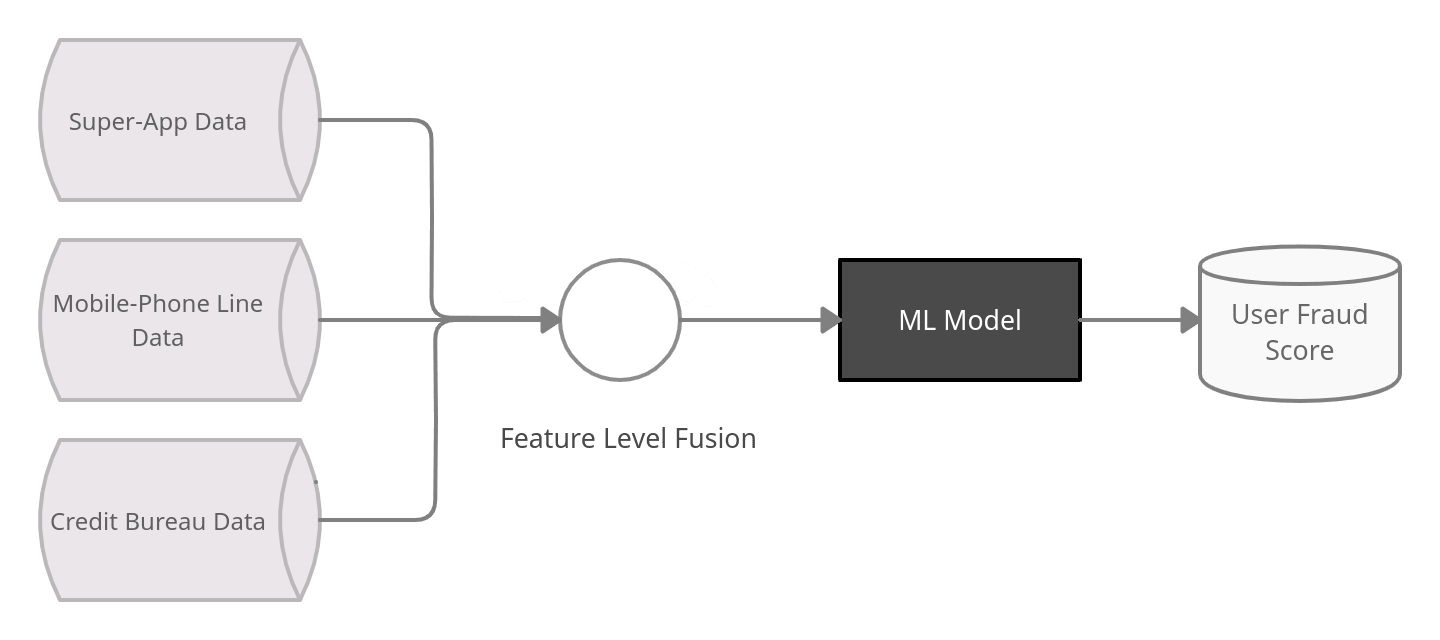}
        \caption{Graphic representation of the proposed approach, where we perform a feature level fusion of the input variables for the proposed scenarios, afterwards we train an Gradient Boosting Decision Tree that outputs a fraud risk Score that is a continuous value between 0 and 1, where 1 is given for fraudsters.}
        \label{fig:flowchart}
\end{figure*}

\subsection{Mobile Phone Line Data}

Mobile-phone data has also been effectively used for the prediction of credit worthiness. Miller \cite{2020owi} points out that the leaders of companies such as \textit{Fidelity Investments}, \textit{Kiva} and \textit{Juvo} discuss how each of them have used mobile data to build alternative credit files for the previously non-banked and underbanked. They also highlight how these data sources can help in the establishment of a financial identity that is more commonly found in a population given the everyday usage of mobile phones. There are also benefits in inclusiveness and reducing credit inequalities when using mobile phone data. Traditional banks in developing markets often reject credit applications from vulnerable groups such as the poor, women, and small businesses because of the lack of a previous credit bureau history \cite{2020da}.

Traditional credit scoring institutions like FICO have also ventured into mining mobile phone data for their models. Such an example is the \textit{FICO Score XD} that was established by FICO in the United States \cite{2017fico}.

Additionally, mobile phone data has also been empirically proven to be effective in the detection of fraudulent behaviour. Óskarsdóttir et al.~\cite{2019bravomobile} demonstrate how including call networks information adds value in terms of profit by applying a profit measure and profit-based feature selection. Ots et al.~\cite{2020ots} also prove that mobile phone usage data can be used to make predictions and find the best classification method for credit scoring.

Therefore, we can use the data taken from the customers' transactional behaviour and their mobile phone consumption patterns to obtain more insights on whether they will commit credit card origination fraud or not. In terms of transaction patterns, we can determine how consistent the user is in the frequency of their payments, as well as how consistent is the amount spent by the user's mobile line in each period or if there is a clear trend or seasonality in their consumption patterns. As for the behavioural features, additional information like the tenure of the mobile phone line, or the amount of calls that are received and sent allows us to better characterize the user's profile.

\section{Approach}

In order to raise the bar against credit card fraudsters we propose a combination of not commonly used data sources in traditional banking institutions such as the super-app data and mobile phone line data. Additional to both of those sets of alternative data, we also train a model using traditional credit bureau data in order to build a model baseline for prediction of credit card origination fraud. Therefore, throughout this paper we aim to answer the following research questions:

\begin{itemize}
    \item \textbf{RQ1:} Is alternative data useful in the prediction of credit origination fraud?
    \item \textbf{RQ2:} Is the mobile phone line data complementary to the super-app and Bureau data for credit origination fraud detection?
    \item \textbf{RQ3:} Which model provides the best financial results for the detection of credit card origination fraud?
\end{itemize}

As we are interested in the gains when using alternative data for credit card fraud detection, we design 6 different sets of inputs for a machine learning model. In Table~\ref{tab:scenarios} we summarize the feature set included in every scenario, where the distinction between scenarios is the sources of data that are included in the input vector. Notice that we perform a feature-level fusion of the input variables for each combination. Afterwards, we train a Gradient Boosting Decision Tree aimed to output the user fraud risk score. This fraud risk score is a continuous value between 0 and 1, where higher values indicate a higher fraud risk. A graphical description of the proposed approach can be found in Figure~\ref{fig:flowchart}.

Then we measure the effectiveness of each of those sets by themselves or combined in the proposed approach via commonly used machine learning metrics such as: ROC Curve AUC, F1-Score, Precision $\&$ Recall; and a custom created financial loss as well. And finally we estimate the importance of the variables for each model using SHAP values \cite{lundberg2017} to give us more insight on how the model is using the different sources of information.


\begin{table}[t]
    \centering
    \caption{Description of input data for each of the 6 scenarios proposed}
    \begin{tabular}{c|c|c}
        \textbf{Scenario} & \textbf{Input Data} & \textbf{Acronym} \\
        \hline\hline
        Scenario 1 & Credit-Bureau Data only & C \\\hline
        Scenario 2 &  Super-App Data only & S \\\hline
        Scenario 3 & Mobile-Phone Data only & M \\\hline
        Scenario 4 & Super-App Data + Mobile-Phone Data & S+M \\\hline
        Scenario 5 & Super-App Data + Credit-Bureau Data & S+C \\\hline
        Scenario 6 & \shortstack{Credit-Bureau Data +  \\ Super-App Data + Mobile-Phone Data} & S+M+C  \\
    \end{tabular}
    \label{tab:scenarios}
\end{table}


\subsection{Financial Loss Function}

We also set up a custom financial loss function, giving different valuation to false positives and false negatives. As previously mentioned, in a super-app environment, the cost incurred with false predictions is not as straightforward as in traditional banks. The False Negative value is indeed similar, being the amount of the line given to an actual fraudulent user. But, the False Positive amount probably means the loss of a client in all of the super app's verticals, not just the loss of what would have been a trustworthy credit user. Therefore, we set up the following financial loss function:

\begin{equation}
    \mathcal{L}=\sum_{i\in FN}ACL + \sum_{i\in FP}CLV \cdot P(ch|r)    
\end{equation}


where, as previously mentioned, the cost of a false negative $ACL$ is the average amount of the line given to the fraudulent user. The cost for the false positive is the average Customer's Life-time Value, $CLV$, multiplied by the probability that the client leaves the super-app after being rejected for a credit line: $P(ch|r)$. These values are calibrated using average credit lines of users approved so far and estimation of lifetime values and churn probabilities for the company.

\section{Evaluation}

\subsection{Dataset Description}

We consider a full dataset of 86,708 users. Where among them we define the fraud label as those users that were manually confirmed as identity theft fraudsters by experts. We end up with 1,421 such cases overall. Additionally, we separate our train and test sets using a back-testing methodology. Making a time split, with the first 60\% credit card applications as the training set, and the resting 40\% of the applications as the test set. This with the idea that with fraudulent users tactics evolving constantly to bypass prediction models, we make sure that future and more complex patterns that may be adopted by the fraudulent users are not available in the data for the train set and used to better detect past fraudulent activity. Table \ref{tab:data_specifications} describes the specifications for the Train and Test datasets.

\begin{table}[h!]
\centering
  \caption{Dataset Description}
  \label{tab:data_specifications}
  \begin{tabular}{c c c} 
    \hline 
    \textbf{Dataset} & \textbf{\# of Users} & \textbf{Fraud Percentage (\%)} \\
    \hline
    \textit{Train} & 60,708      & 2.18\% \\
    \textit{Test} & 26,018      & 0.37\% \\
    \hline
  \end{tabular}
\end{table}

Moreover, for each user in the dataset we consider three sets of features:

\begin{itemize}
    \item \textbf{Super-App Data}
    
    Where we have 8 different features. Describing the user's transactionality: \% of amount spent in restaurants, \% of amount spent in supermarkets, between others. And demographic variables, like: User's tenure in the digital platform, the number of registered devices in the platform, etc.
    
    \item \textbf{Mobile-Phone Data}
    
    Where we add 6 features. These describe the users Mobile Phone Line's transactionality, with variables such as: the cadence, measuring how consistent the user payments are in amount, and the amount of phone calls received and sent. In terms of demographic variables we consider features such as: the tenure of the Mobile Phone Line, and whether the user has a pre-paid or post-paid line.
    
    \item \textbf{Credit-Bureau Data}
    
    That contains 9 different variables. And amongst them are features such as:  the credit score, and the accumulated credit lines the user has.
\end{itemize}

\subsection{Experimental Setup}

For the comparison, we train a Gradient Boosting Decision Tree~\cite{chen2016xgboost} classification model which hyper-parameters where tuned for optimal performance via a grid search methodology, using the train set of credit card applications with each of scenarios described above. Then, we test the model on the previously mentioned test set.

\subsection{Results}

\begin{table*}[t]
\centering
  \caption{Model Results Comparison ML Metrics}
  \label{tab:res}
  \resizebox{\textwidth}{!}{
  \begin{tabular}{p{\dimexpr.30\linewidth-1\tabcolsep}p{\dimexpr.15\linewidth-1\tabcolsep}p{\dimexpr.15\linewidth-1\tabcolsep}p{\dimexpr.15\linewidth-1\tabcolsep}p{\dimexpr.15\linewidth-1\tabcolsep}p{\dimexpr.15\linewidth-1\tabcolsep}}
    \hline
    \textbf{Model} & \textbf{AUC (\%)} & \textbf{F1-Score (\%)} & \textbf{Precision (\%)} & \textbf{Recall (\%)} \\
    \hline
    Credit Bureau & 72.62$\pm$0.31 & 8.39$\pm$0.30 & 9.03$\pm$0.92 & 11.77$\pm$1.19\\
    Super-App & 79.46$\pm$0.20 & 18.99$\pm$0.39 & 33.31$\pm$2.05 & 14.05$\pm$0.42\\
    Mobile Phone & 68.24$\pm$0.40 & 16.98$\pm$0.32 & 21.00$\pm$1.09 & 14.96$\pm$0.38\\
    Super-App + Mobile Phone & 78.46$\pm$0.35 & 27.75$\pm$0.38 & 31.16$\pm$1.05 & 25.93$\pm$0.69\\
    Super-App + Credit Bureau & \textbf{82.24$\pm$0.33} & 32.88$\pm$0.31 & 35.11$\pm$1.28 & 32.27$\pm$0.83\\
    Super-App + Mobile Phone + Credit Bureau & 81.05$\pm$0.33 & \textbf{38.66$\pm$0.34} & \textbf{40.58$\pm$0.83} & \textbf{37.48$\pm$0.62}\\
    \hline
  \end{tabular}}
\end{table*}

\begin{figure}
        \centering
        \includegraphics[width=0.5\textwidth]{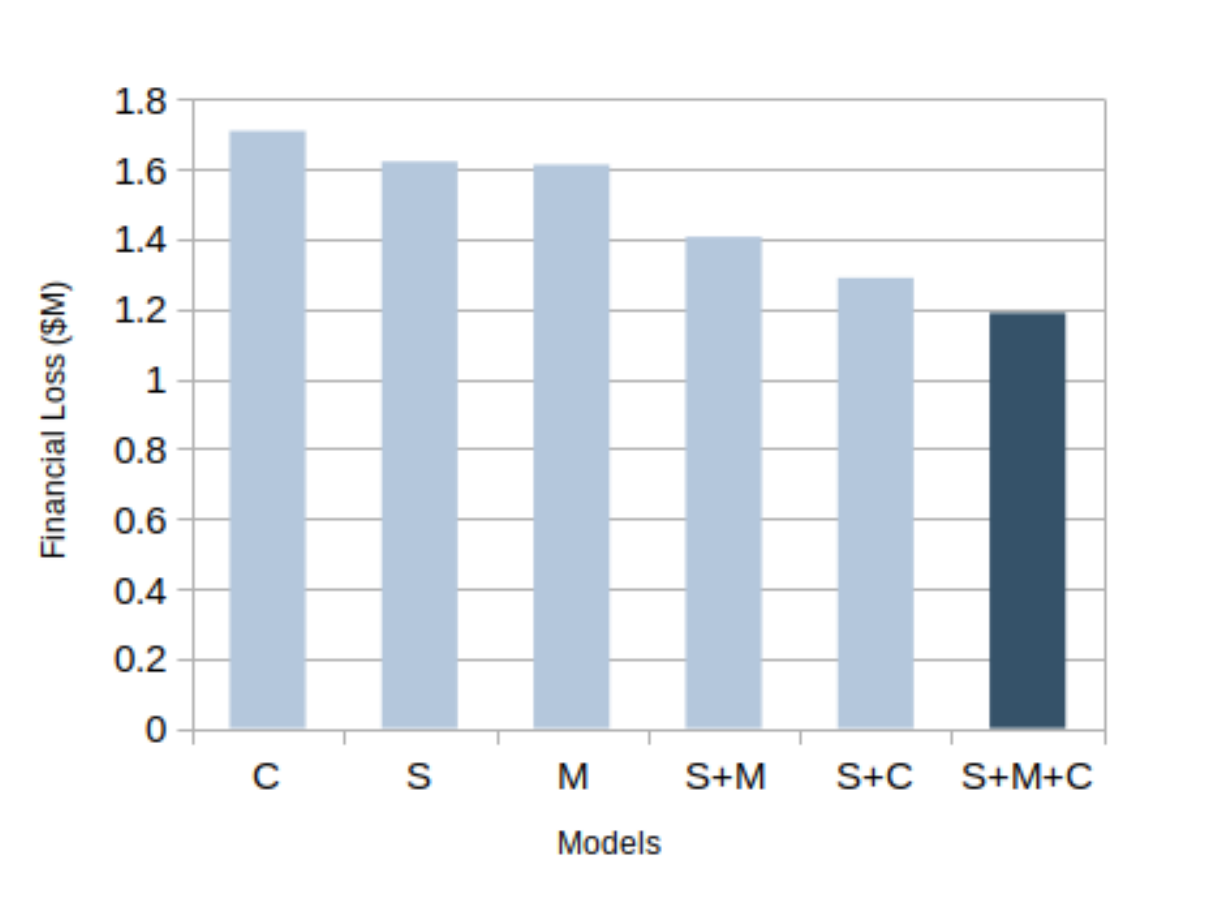}
        \caption{Comparison of the custom created  financial loss for the 6 considered scenarios, C stands for Credit Bureau data, S stands for Super-App data, and M stands for Mobile Phone Line data }
        \label{fig:finloss}
\end{figure}

All of these models are compared using: 1. ROC-Curve area under the curve (AUC), 2. Precision $\&$ Recall scores utilizing the optimal threshold for the F-1 Score, 3. The F-1 Score, and 4. the actual financial loss incurred by the model output. There results are compared for 100 different bootstrap runs, each with a different seed, to have a more robust comparison of the effectiveness of the model. Additionally, we analyze the results with the use of Shap Values for the determination of the feature importance in each of the models, these values are a game theoretic approach that estimates an importance value to each feature for a particular prediction \cite{lundberg2017}.

Table \ref{tab:res} summarizes the results for the scenarios proposed in terms of traditional machine learning metrics. And in Figure \ref{fig:finloss} we can see a graphic representation about how each model compares in terms of the custom financial loss function.

We can see that evaluating every feature set by their own, the super-app data has the best performing AUC and F1-Score in terms of credit card origination fraud prediction. Although, it is slightly improved by the Mobile-Phone data in terms of the financial loss incurred with those predictions, this can be due to the Mobile Phone's data model having a better recall. The Credit Bureau data model performs the worst out of the three models for all metrics except for the AUC.

When combining the super-app data features with just one more dataset, we can see that the combination with the Credit Bureau data is more effective than the combination with the Mobile Phone Data for the fraudulent user's prediction. This iteration of the model is better by all of the chosen evaluation metrics.

Additionally, we can see that the complete model is improved by the Super-App + Credit Bureau model in terms of AUC. But, it is improved significantly in every single metric other than the AUC, where the complete model was the best performing of all the iterations.

Figure~\ref{fig:shap} shows the mean absolute SHAP values for the features with the highest importance in the final model. Each bar represents the mean absolute SHAP value for the corresponding feature across all observations in the test set. We observe that two out of the three most important feature (those with the highest mean average SHAP values) come from the phone-line usage data. Phone line tenure, which measures the amount of time a user has with their current phone line, has a significant contribution of 0.75 absolute SHAP value on average on the model's predicted fraud score. Similarly, a Risk score calculated using a larger set of phone-line usage has a great impact on the model's output. The most important feature, however, is Tenure, which measures the amount of time a customer has been a super-app customer. 

\begin{figure}[t]
    \begin{minipage}{0.45\textwidth}
        \includegraphics[width=\textwidth]{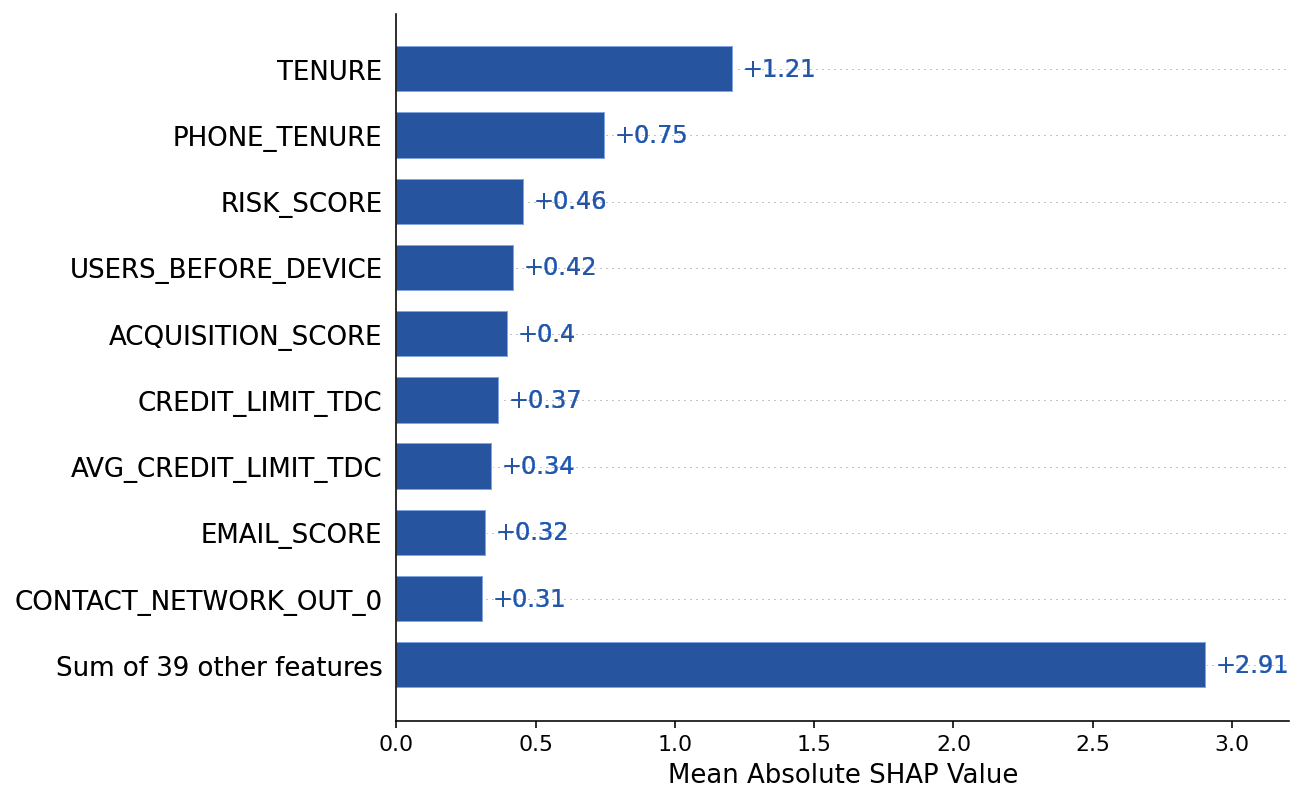}
        \caption{Mean absolute SHAP values for model with all variables. The 48 variables come from the encodings used for the before-mentioned features in the Dataset Description.}
        \label{fig:shap}
    \end{minipage}\hfill
\end{figure}

For the rest of the top nine features, we can see that they are evenly distributed among the 3 feature sets. Two more come from the super-app data (Acquisition Score and Email Score), two come from the credit bureau (Credit Card Credit Limit and Average Credit Card Limit), and one more comes from the phone-line usage data (Contact Network Out, which measures the amount of outgoing contacts a user has).

\subsection{Discussion}

Notice in Table \ref{tab:data_specifications} that the fraud percentages go down in the test set compared to the train set. This is due to the time split of the data, where the training set corresponds to the first $60\%$ of the data in chronological order and the test set is the resulting $40\%$. This drop in fraud can reflect two things. It may reflect that fraud-detection models are evolving over time and incorporating new information, which increases the barriers for the fraudsters and leads to an overall fraud reduction. However, this drop can also reflect that fraudsters' strategies evolve over time faster than detection systems, and a larger fraction of fraudsters are going undetected by experts. This train-test split, we think, is a fair evaluation for our approach, as it resembles how the model would behave in a productive setting.

The main limitation of our results is that the improvements we are obtaining by using alternative data are specific for the population that appear in each of our three data sources: credit bureau, super-app, and mobile phone company. Given that appearing in each of the sources of data may reflect specific characteristics of the users, who self select into using the super-app, acquiring a phone line with a specific company, and having financial products that makes them appear in bureau, we cannot extrapolate the results to the whole population. Further research should be carried out to understand whether alternative data sources work well for different segments of the population.

Additionally we have to also consider the ethical implications of this study, while the use of these datasets is proven to very well improve the detection of fraudsters; it might also carry the risk violating people's privacy. And as such, this balance has to be taken into account.

Finally, our results may be a lower bound on the improvements that can be attained by using super-app and phone-line data, given that in this paper we are only exploring a feature-level fusion. For example, ensembles of models that are trained using different sources of data, or a careful feature engineering that combines information from different data sources, may generate even larger gains.

\section{Conclusion}
In this work we have presented a comprehensive comparison of the performance of models aimed to detect credit card fraud for different sources of data as the input. In particular, we compared 6 combinations of alternative data features as the model input using a feature-level fusion strategy. We showed that best performance is achieved when including alternative data from super-apps and telecommunication data for detecting credit card fraud. Our evaluation was performed over a realistic dataset that contain data for over 90,000 users worldwide. In the future, we would like to expand our set of alternative data to a wider spectrum, for instance, by adding data of studies and job profiling.






\bibliographystyle{IEEEtran}
\bibliography{references.bib}

%



\end{document}